\documentclass[conference]{IEEEtran}

\makeatletter

\def\ps@IEEEtitlepagestyle{%
  \def\@oddfoot{\mycopyrightnotice}%
  \def\@evenfoot{}%
}
\def\mycopyrightnotice{%
  {\footnotesize XXX-X-XXXX-XXXX-X/XX/\$XX.00~\copyright~20XX IEEE\hfill}
  \gdef\mycopyrightnotice{}
}

\IEEEoverridecommandlockouts
\usepackage{cite}
\usepackage{amsmath,amssymb,amsfonts}
\usepackage{algorithmic}
\usepackage{graphicx}
\usepackage{textcomp}
\usepackage{xcolor}
\usepackage{url}
\def\BibTeX{{\rm B\kern-.05em{\sc i\kern-.025em b}\kern-.08em
    T\kern-.1667em\lower.7ex\hbox{E}\kern-.125emX}}

\usepackage{eso-pic}
\begin{document}

\title{\vspace*{1cm} Unsupervised Clustering of Public Transport Vehicles Using Deep Morphological Features and Color Histograms}

\author{\IEEEauthorblockN{Juan Irving Vasquez}
\IEEEauthorblockA{\textit{Centro de Innovación y Desarrollo Tecnológico en Cómputo} \\
\textit{Instituto Politécnico Nacional}\\
Mexico City, Mexico \\
jvasquezg@ipn.mx}
}

\maketitle

\begin{abstract}
Surveillance cameras along urban roads produce large volumes of unlabeled video that could support the monitoring of public transport, but grouping the observed vehicles into meaningful categories without manual annotation remains challenging, since buses from different routes often share the same body structure and differ mainly in their color schemes. This study presents an unsupervised clustering methodology for public transport vehicles that combines morphological features extracted with a frozen DINOv2 Vision Transformer and handcrafted HSV color histograms through weighted early fusion, followed by UMAP projection and K-Means clustering. The method was evaluated on 7,972 bus images extracted from about 1,460 hours of video of an arterial road in Mexico City, with the goal of isolating the buses of the city's public bus network (RTP) without supervision. With a fusion of 65\% morphological and 35\% color features, the pipeline produced 14 clusters that separate distinct routes and vehicle types, and it increased the precision of the target cluster with respect to morphological features alone. These results show that color complements self-supervised features for route-level clustering, provided that it does not dominate the representation. The dataset is available under MIT license at: \url{https://www.kaggle.com/datasets/irvingvasquez/public-bus-images}
\end{abstract}


\section{Introduction}
\label{sec:introduction}
 
Public transport is the backbone of mobility in large cities, and its efficient management requires continuous information about which vehicles operate on each corridor, how often they pass, and whether they follow their assigned routes. Intelligent Transportation Systems (ITS) increasingly obtain this information from the surveillance cameras already installed along urban roads. These cameras generate thousands of hours of video, far more than can be inspected or annotated manually, so methods that extract useful information from such footage without human supervision are of great practical value.
 
In this work, we address the problem of grouping public transport vehicles captured by a fixed surveillance camera into semantically meaningful categories, such as transit routes and operators, without using labeled data. The problem is challenging for two complementary reasons. First, buses operating on different routes often share the same body structure and differ mainly in their paint schemes. Second, vehicles of different types, such as full-size buses, \textit{microbuses} (small buses), and vans, may share similar colors while differing substantially in shape. A useful representation must therefore capture both morphology and color. 

 
Most vision-based vehicle classification systems are supervised and rely on large, site-specific annotated datasets with fixed taxonomies in which ``bus'' is a single class~\cite{berwo2023deep,butt2021convolutional,neupane2022realtime}. Self-supervised learning has reduced the need for labels during representation learning~\cite{uelwer2025survey}, but in the vehicle domain it has mainly been used as pre-training for a supervised classifier~\cite{ma2023imagebased}. Unsupervised clustering, on the other hand, has been applied mainly to vehicle re-identification, which groups images of the same physical vehicle rather than different vehicles that share a category~\cite{qiu2024camera}. Self-supervised Vision Transformers such as DINOv2~\cite{oquab2024dinov2} now provide general-purpose descriptors that capture image structure well without fine-tuning. However, their features may under-represent pure chromatic information, which is crucial for distinguishing transit routes.
 
\begin{figure*}[tb]
\centering
\includegraphics[width=\textwidth]{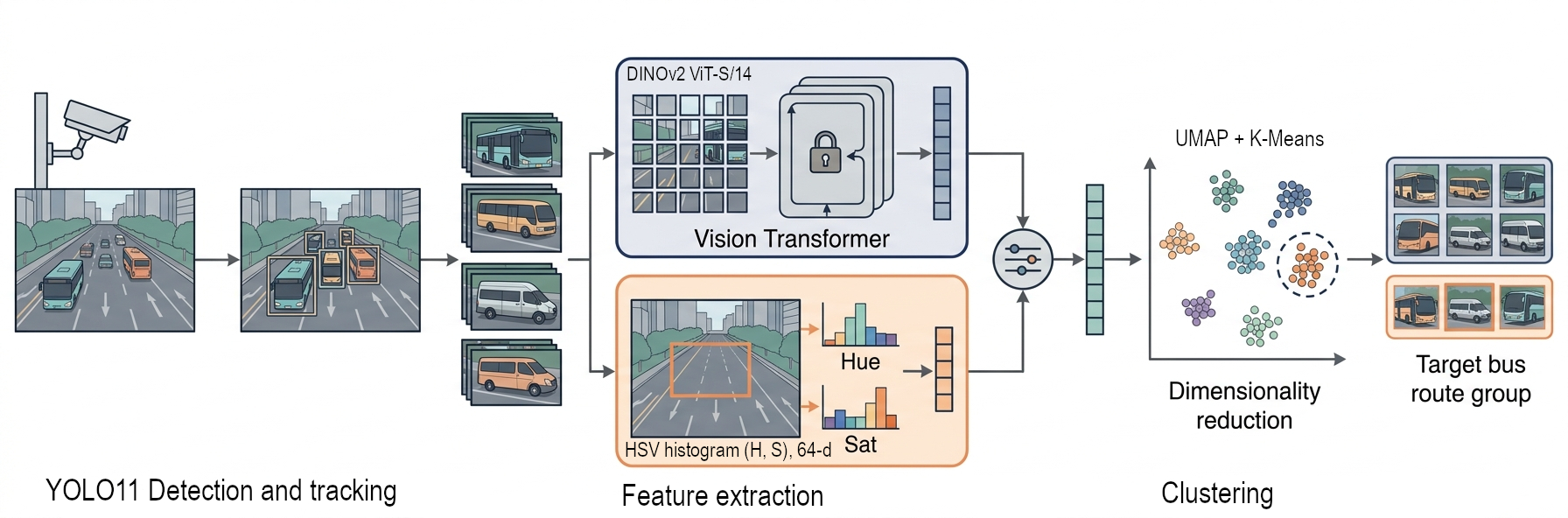}
\caption{Overview of the proposed unsupervised clustering pipeline. Buses are detected and tracked in surveillance video, each crop is described by frozen DINOv2 morphological features and HSV color histograms, both descriptors are combined through weighted early fusion, and the fused vectors are projected with UMAP and grouped with K-Means. Bus illustrations were generated with AI.}
\label{fig:methodology}
\end{figure*}

To address this problem, we propose a hybrid unsupervised clustering pipeline for public transport vehicles. Buses are first detected and tracked in the video, and a single crop is stored per vehicle. Each crop is then described by two complementary descriptors: a morphological descriptor given by the class token of a frozen DINOv2 model, and a color descriptor given by hue and saturation histograms computed in the HSV color space. Both descriptors are normalized and combined through weighted early fusion, and the fused vectors are projected with UMAP and grouped with K-Means, as illustrated in Fig.~\ref{fig:methodology}. We apply the pipeline to images of buses operating on one of the main arterial roads of Mexico City, and evaluate whether it can isolate the buses of the \textit{Red de Transporte de Pasajeros} (RTP), the city's public bus network, as a coherent cluster.

The main contributions of this paper are as follows:
\begin{itemize}
    \item An unsupervised pipeline that fuses frozen self-supervised morphological features with handcrafted color histograms to cluster public transport vehicles captured by a surveillance camera, without requiring labels at any stage of feature extraction or clustering.
    \item A dataset of 7,972 bus images extracted from about 1,460 hours of video of a Mexico City arterial road, manually annotated for evaluation and publicly available at \url{https://www.kaggle.com/datasets/irvingvasquez/public-bus-images}.
    \item An ablation study on the fusion weight showing that a moderate color contribution increases the precision of the target cluster with respect to morphological features alone, whereas an excessive color weight leads to over-segmentation or to the collapse of the clustering space.
\end{itemize}
 
The remainder of this paper is organized as follows. Section~\ref{sec:related_work} reviews related work on vehicle classification, self-supervised learning, and clustering. Section~\ref{sec:methods} describes the dataset and the proposed clustering pipeline. Section~\ref{sec:results} presents the quantitative and qualitative results. Finally, Section~\ref{sec:conclusions} concludes the paper and outlines future work.

\section{Related Work}
\label{sec:related_work}

Vision-based vehicle classification has evolved from hand-crafted descriptors, such as Haar-like features and histograms of oriented gradients (HOG), combined with shallow classifiers such as support vector machines and AdaBoost, to convolutional neural networks trained end-to-end~\cite{berwo2023deep}. Supervised deep models now reach high accuracy. Butt \emph{et al.}~\cite{butt2021convolutional} fine-tuned several CNNs on 10,000 images of six vehicle types captured under adverse illumination and reported 99.68\% accuracy, and Neupane \emph{et al.}~\cite{neupane2022realtime} fine-tuned YOLO detectors on nearly 30,000 manually annotated samples of seven classes, including buses. The latter study also shows the cost of domain shift: accuracy roughly doubled (from at most 30\% to 71\%) only after fine-tuning on data from the target cameras. These systems therefore depend on large, site-specific annotated datasets and on fixed taxonomies in which ``bus'' is a single class, so they cannot distinguish buses of different routes or operators.

Self-supervised learning (SSL) reduces this dependence by learning representations from unlabeled images~\cite{uelwer2025survey}. In the vehicle domain, Ma and Yang~\cite{ma2023imagebased} showed that Vision Transformer features learned with DINO and data2vec, combined with wheel-position cues, reach 97.2\% top-1 accuracy on the 13 axle-based classes of the U.S. Federal Highway Administration. However, SSL is used there only as pre-training for a supervised classifier, and buses again form a single class.

Deep clustering removes labels altogether by coupling a representation-learning module with a clustering module~\cite{zhou2024comprehensive}. Zhou \emph{et al.}~\cite{zhou2024comprehensive} group these methods into multistage, generative, iterative, and simultaneous approaches; our pipeline belongs to the multistage family, in which features are extracted first and clustered afterwards. Hou \emph{et al.}~\cite{hou2026comprehensive} note that, before deep learning, image clustering relied on hand-crafted descriptors such as color histograms, and identify robustness to noise and real-world applicability as open challenges for deep image clustering. In the vehicle domain, clustering has mainly been used for unsupervised re-identification (Re-ID). Qiu \emph{et al.}~\cite{qiu2024camera}, for instance, generate pseudo-identity labels with a camera-aware DBSCAN scheme and refine them with a focal contrastive loss, outperforming previous unsupervised methods on VeRi-776 and VERI-Wild. Re-ID, however, groups images of the \emph{same physical vehicle}, whereas our goal is to group \emph{different} vehicles that share a semantic category.

In summary, existing work either classifies vehicles into coarse, predefined types~\cite{butt2021convolutional,neupane2022realtime,ma2023imagebased} or matches individual vehicles across cameras~\cite{qiu2024camera}. To our knowledge, the unsupervised, category-level partitioning of the bus class, in which vehicles of different routes share a body structure and differ mainly in livery, has not been addressed. Moreover, none of the reviewed works combines self-supervised deep features with explicit color descriptors, even though color is a decisive cue for route identity. This work addresses both gaps.

\section{Materials and Methods}
\label{sec:methods}

\subsection{Dataset}

The dataset was constructed from approximately 1,460 hours of continuous video recorded during June and July 2026. Recordings were acquired with a single fixed camera mounted at a height of 5~m and positioned 30~m from Circuito Interior, one of the main arterial roads of Mexico City. The camera was oriented toward the roadway, providing a consistent view of passing traffic, including public transport buses operating on multiple routes.

\subsubsection{Bus Detection}

\begin{figure}[tb]
\centering
    \includegraphics[width=0.32\columnwidth,height=0.24\columnwidth,keepaspectratio=false]{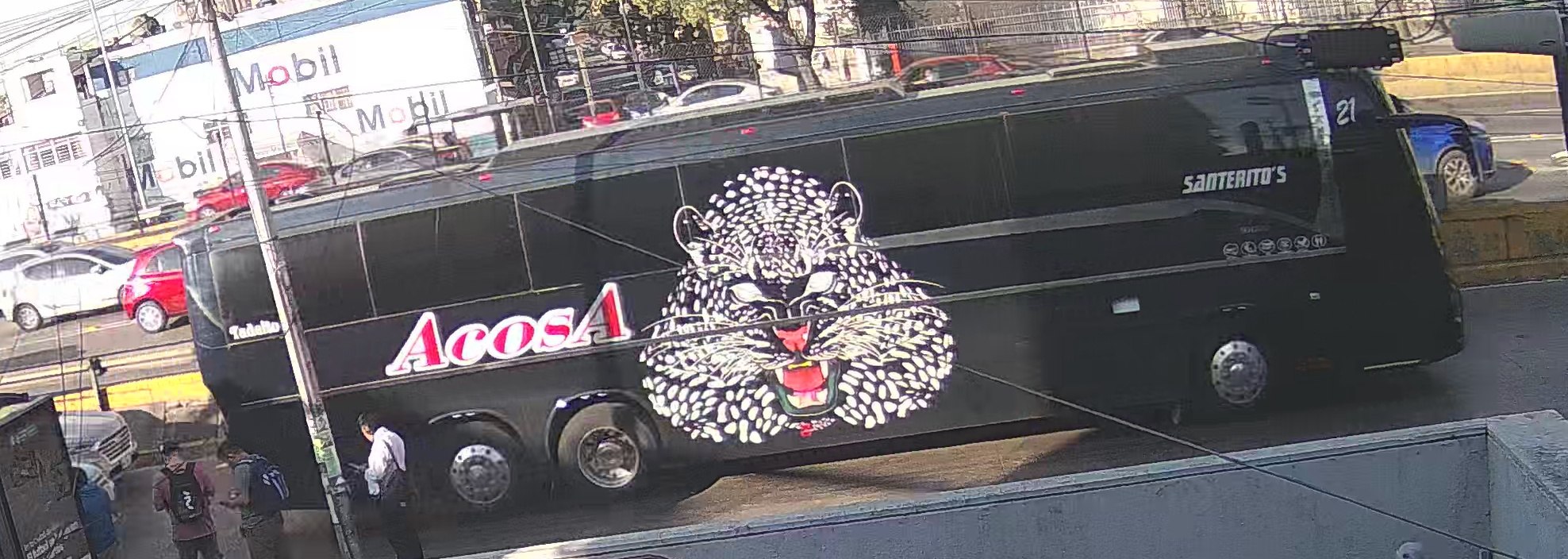}\hfill
    \includegraphics[width=0.32\columnwidth,height=0.24\columnwidth,keepaspectratio=false]{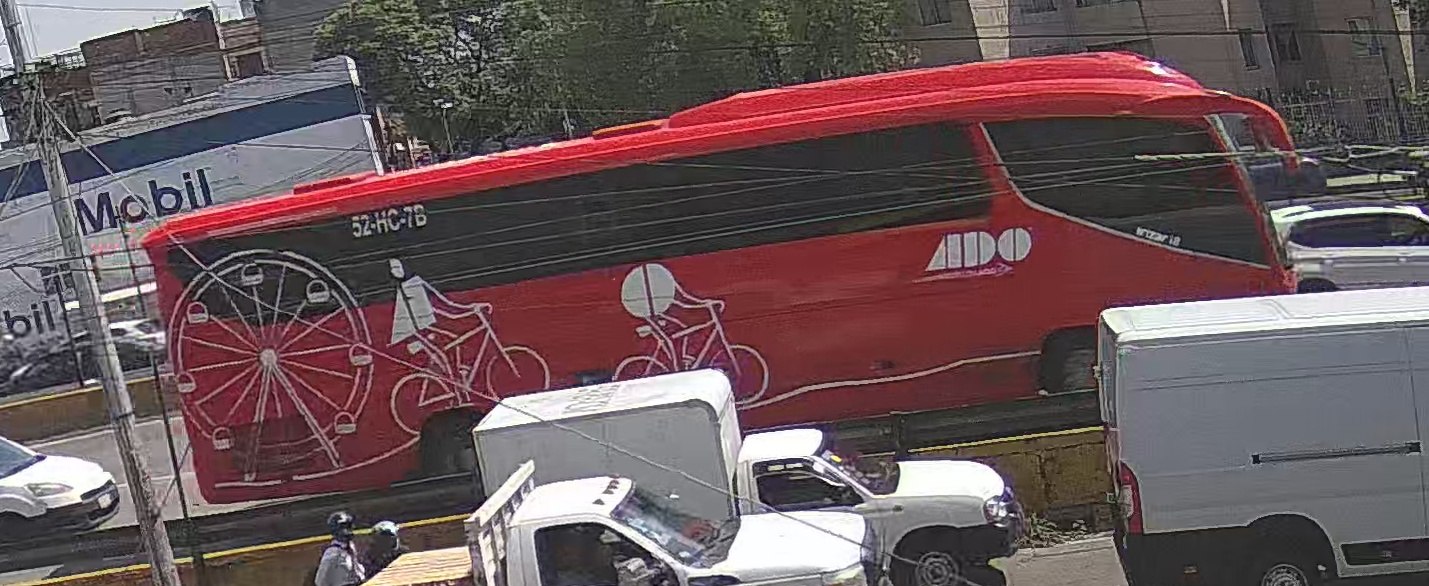}\hfill
    \includegraphics[width=0.32\columnwidth,height=0.24\columnwidth,keepaspectratio=false]{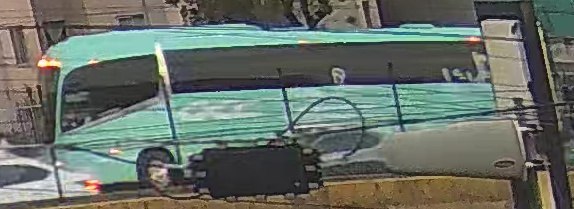}\\[4pt]
    \includegraphics[width=0.32\columnwidth,height=0.24\columnwidth,keepaspectratio=false]{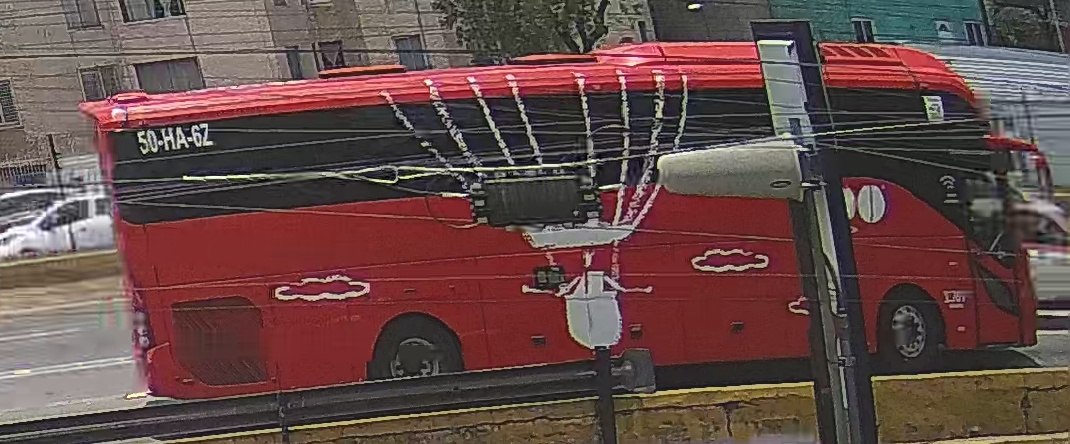}\hfill
    \includegraphics[width=0.32\columnwidth,height=0.24\columnwidth,keepaspectratio=false]{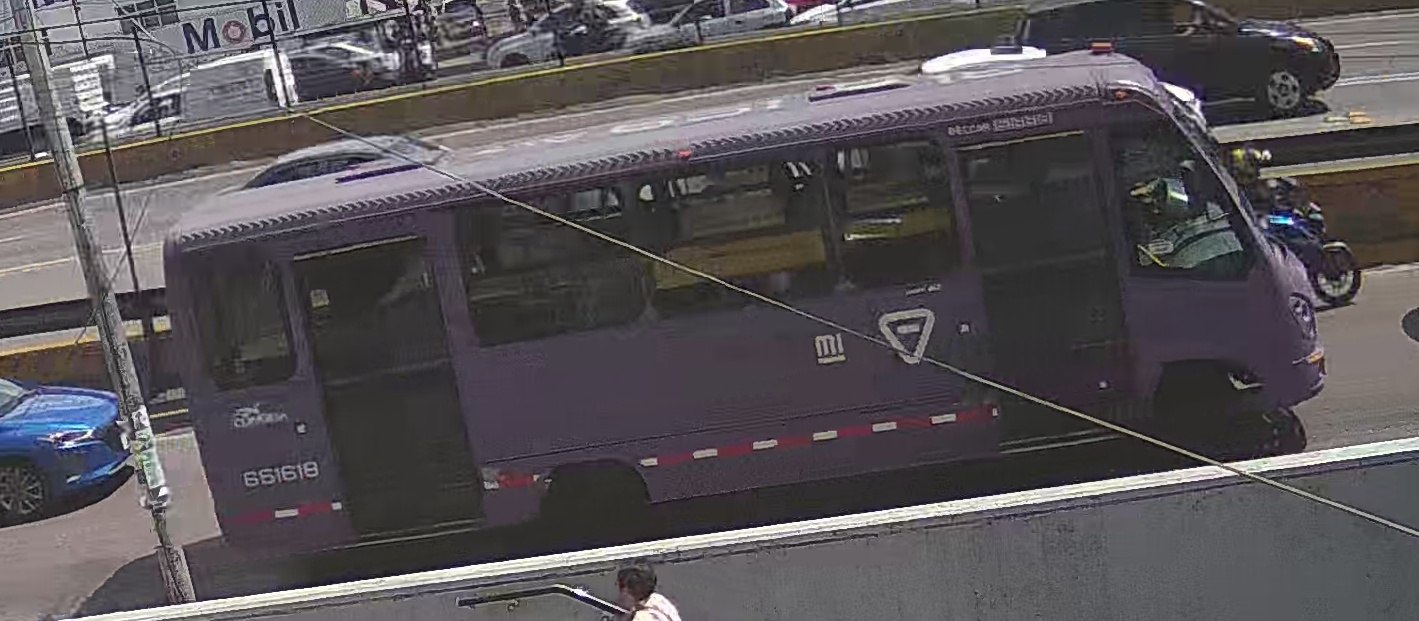}\hfill
    \includegraphics[width=0.32\columnwidth,height=0.24\columnwidth,keepaspectratio=false]{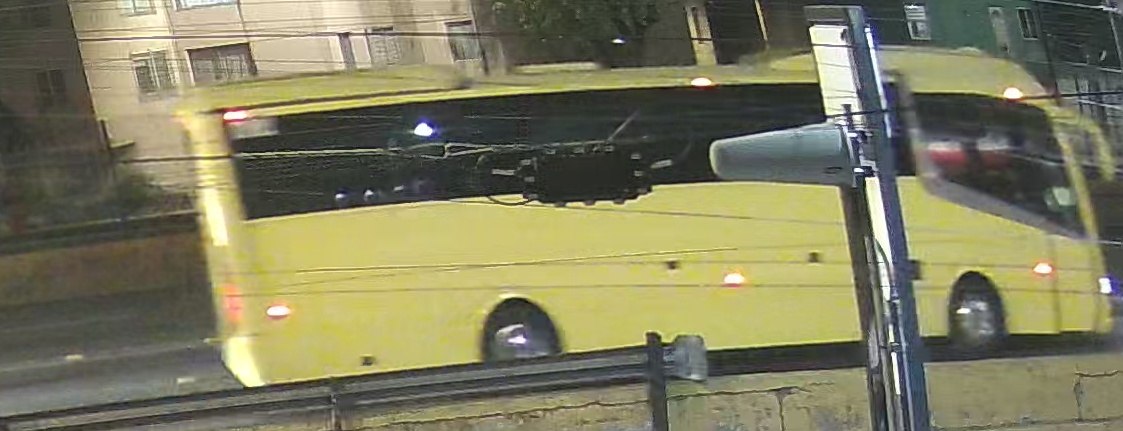}\\[4pt]
    \includegraphics[width=0.32\columnwidth,height=0.24\columnwidth,keepaspectratio=false]{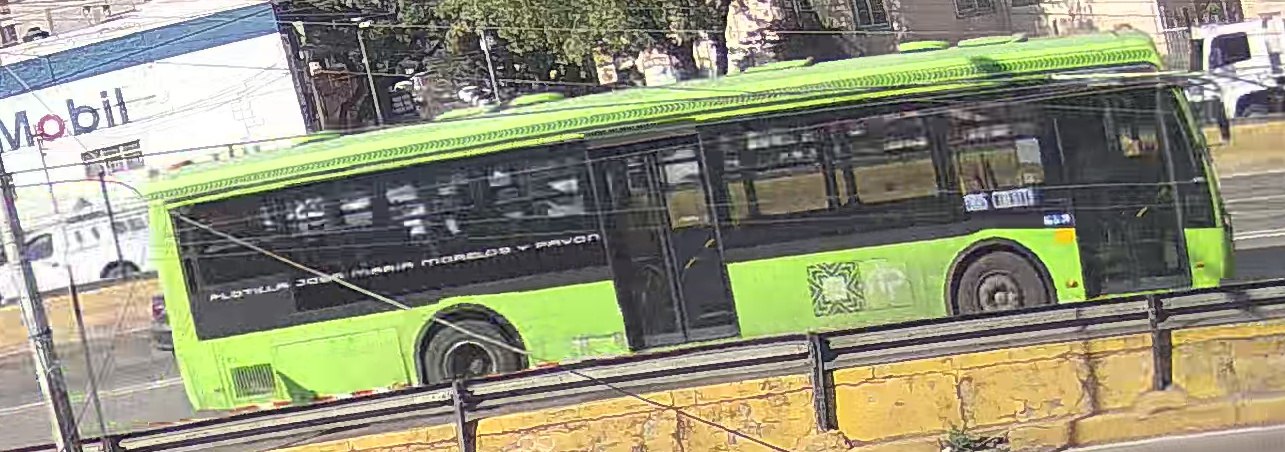}\hfill
    \includegraphics[width=0.32\columnwidth,height=0.24\columnwidth,keepaspectratio=false]{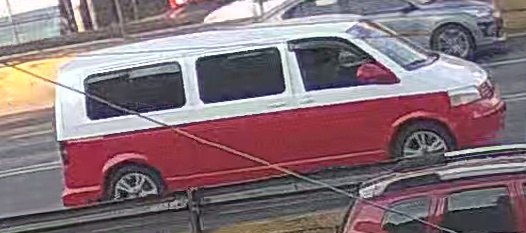}\hfill
    \includegraphics[width=0.32\columnwidth,height=0.24\columnwidth,keepaspectratio=false]{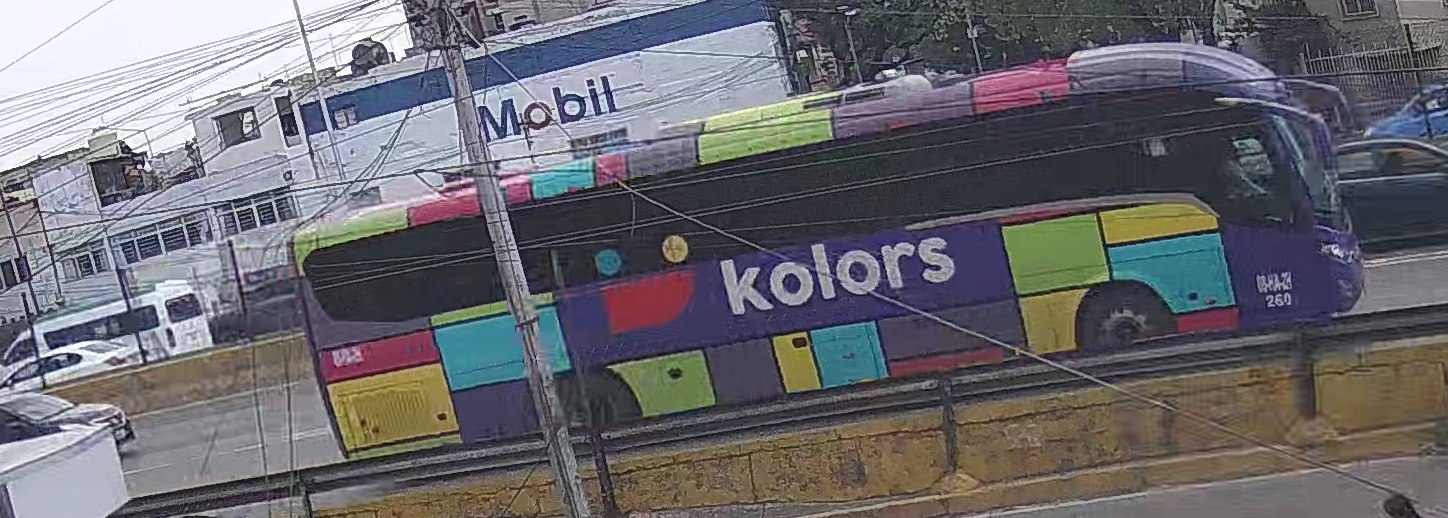}\\[4pt]
    \includegraphics[width=0.32\columnwidth,height=0.24\columnwidth,keepaspectratio=false]{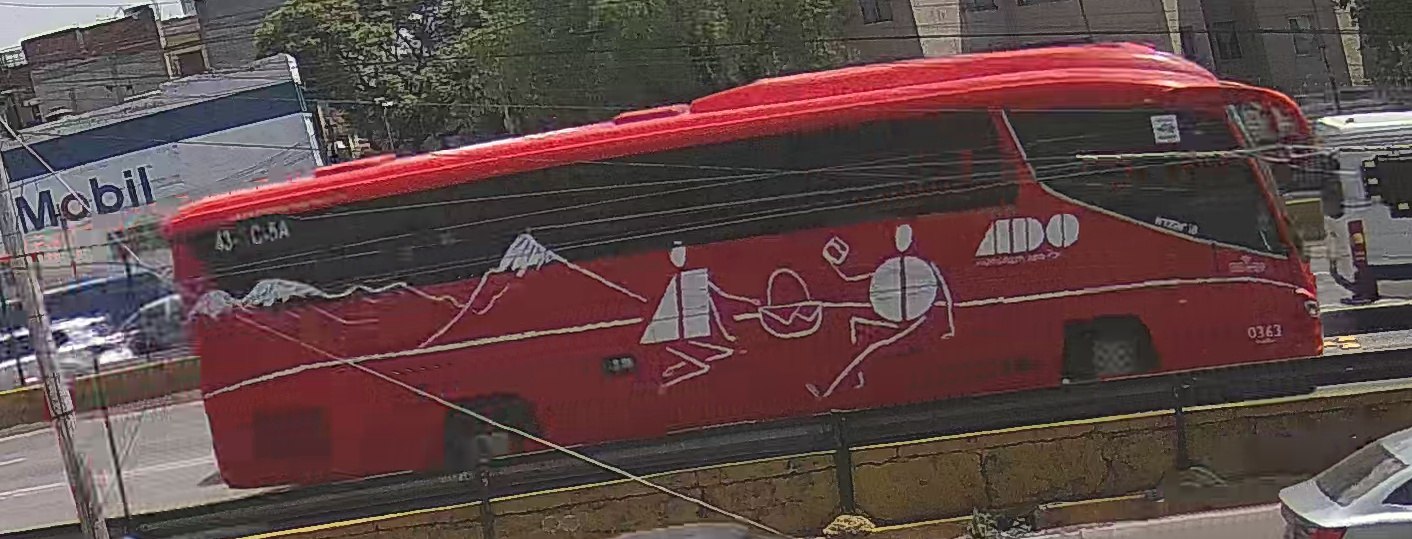}\hfill
    \includegraphics[width=0.32\columnwidth,height=0.24\columnwidth,keepaspectratio=false]{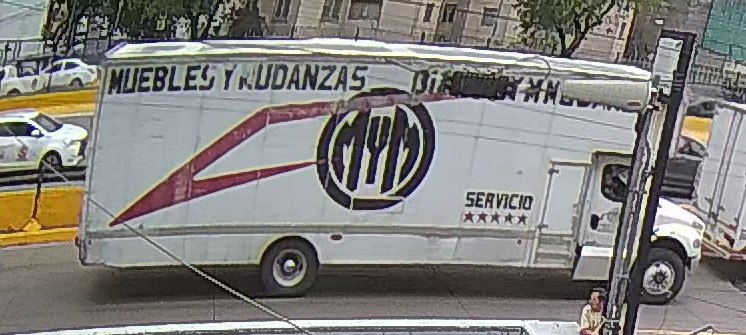}\hfill
    \includegraphics[width=0.32\columnwidth,height=0.24\columnwidth,keepaspectratio=false]{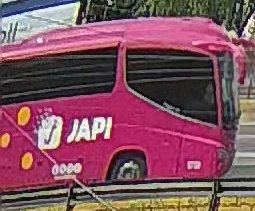}
\caption{Example images from the dataset, cropped from the surveillance video after bus detection and tracking. Because YOLO26 does not distinguish between bus subtypes, the dataset includes full-size buses, microbuses, and vans.}
\label{fig:dataset_examples}
\end{figure}

Given the large volume of recorded video, manual annotation of the full videos was impractical. The videos were therefore preprocessed using the YOLO26 object detector~\cite{ultralytics_yolo26} combined with a multi-object tracker. Only detections belonging to the \textit{bus} class were retained, and a single image was stored for each tracked vehicle to avoid redundant samples of the same bus. To prevent truncation of the vehicle body, each bounding box was expanded by 10\% before cropping. This process yielded a total of 7,972 cropped images of public transport vehicles. Because the detector does not distinguish between subtypes, this set includes buses, microbuses, and vans, all of which were classified as \textit{bus} by YOLO26. Fig.~\ref{fig:dataset_examples} shows examples of the resulting images.

\subsubsection{Ground-Truth Annotation}

Although the proposed clustering approach is fully unsupervised, ground-truth labels were required to evaluate its performance. To this end, all 7,972 images were manually annotated. These labels were used exclusively for evaluation and were not involved at any stage of feature extraction or clustering. Two hierarchical binary labels were assigned to each image: \textit{is bus} and \textit{is RTP bus}, where the latter is a subset of the former. The \textit{is bus} label indicates whether the crop contains a full-size bus and was introduced to assess the precision of the YOLO26-based preprocessing stage, given that smaller vehicles such as microbuses and vans were also detected under the \textit{bus} class. The \textit{is RTP bus} label identifies vehicles belonging to the \textit{Red de Transporte de Pasajeros} (RTP), Mexico City's public bus network, which constitute the target class of this study.

\subsection{Bus Clustering}

The objective of this stage is to group the detected vehicles into clusters that correspond to semantically meaningful categories, such as bus routes and vehicle types, without relying on labeled data. The proposed pipeline consists of three steps: (i) extraction of complementary morphological and color descriptors from each image, (ii) early fusion of both descriptors into a single weighted feature vector, and (iii) unsupervised clustering of the fused representations. The following subsections describe each step in detail.

\subsubsection{Feature Extraction}

Buses operating on different routes often share similar body structures but differ in their paint schemes, whereas vehicles of different types (e.g., full-size buses, microbuses, and vans) may exhibit similar colors while differing substantially in shape. To capture both sources of variation, two complementary types of features were extracted from each image: deep morphological features and handcrafted color features.

\textit{Morphological features:} These features were extracted using DINOv2~\cite{oquab2024dinov2}, a family of Vision Transformer (ViT)~\cite{dosovitskiy2021vit} models trained through self-supervised learning. Unlike text-guided approaches such as CLIP~\cite{radford2021clip}, which depend on paired image and caption data, DINOv2 learns visual representations from images alone. Its training objective combines an image-level self-distillation loss, computed on the class token between a student and a teacher network observing different crops of the same image, with a patch-level masked image modeling loss inherited from iBOT~\cite{zhou2022ibot}.

A key property of DINOv2 is that its features perform well as frozen, general-purpose descriptors, without requiring task-specific fine-tuning. Its authors report strong results with simple linear classifiers and nearest-neighbor retrieval across a wide range of benchmarks, including fine-grained vehicle recognition on Stanford Cars~\cite{oquab2024dinov2}. These properties make DINOv2 well suited to an unsupervised setting such as ours, in which no labeled data are available for fine-tuning and the cluster structure depends directly on the geometry of the embedding space.

In this work, we employed the publicly available \texttt{facebook/dinov2-small} checkpoint, which corresponds to the ViT-S/14 variant of DINOv2 distilled from the ViT-g/14 teacher. Its main architectural characteristics are summarized in Table~\ref{tab:dinov2}. The model was used exclusively as a frozen feature extractor: it was loaded in evaluation mode, gradient computation was disabled, and no fine-tuning or retraining was performed. For each image $I_i$, the [CLS] token of the final hidden layer was taken as a global descriptor of the vehicle's structure, yielding a morphological feature vector
\begin{equation}
    \mathbf{f}^{\text{morph}}_i \in \mathbb{R}^{384}.
\end{equation}

\begin{table}[tb]
\caption{Architecture of the DINOv2 ViT-S/14 checkpoint (\texttt{facebook/dinov2-small}) used as the morphological feature extractor.}
\label{tab:dinov2}
\centering
\begin{tabular}{ll}
\hline
\textbf{Property} & \textbf{Value} \\
\hline
Architecture            & Vision Transformer (ViT-S/14) \\
Transformer blocks      & 12 \\
Attention heads         & 6 \\
Embedding dimension     & 384 \\
Feed-forward network    & MLP with GELU, ratio 4 (1536) \\
Patch size              & $14 \times 14$ \\
Register tokens         & None \\
Parameters              & $\approx$22~M \\
Pretraining data        & LVD-142M (self-supervised) \\
Training regime         & Distilled from ViT-g/14 \\
Weight precision        & float32 \\
\hline
\end{tabular}
\end{table}

\textit{Color features:} To capture the route-specific paint schemes that distinguish otherwise structurally similar buses, color histograms were computed in the HSV color space. HSV was preferred over RGB because it separates chromatic information from intensity, making the descriptor less sensitive to the illumination changes that occur throughout the day in outdoor recordings. Accordingly, only the Hue and Saturation channels were used, and the Value channel was discarded.

Because the expanded bounding boxes include surrounding background such as asphalt, sky, and other vehicles, a central crop covering 50\% of the image area was applied before computing the histograms, so that the descriptor predominantly reflects the vehicle body. One-dimensional histograms with 32 bins were then computed for the Hue and Saturation channels and concatenated, yielding a color feature vector
\begin{equation}
    \mathbf{f}^{\text{color}}_i =
    \left[\, \mathbf{h}^{H}_i \,\|\, \mathbf{h}^{S}_i \,\right]
    \in \mathbb{R}^{64},
\end{equation}
where $\mathbf{h}^{H}_i$ and $\mathbf{h}^{S}_i$ denote the Hue and Saturation histograms of image $I_i$, and $\|$ denotes concatenation.

\subsubsection{Early Feature Fusion}

The morphological and color descriptors differ in both dimensionality (384 versus 64) and numerical scale. To make them comparable, each vector was first L2-normalized independently:
\begin{equation}
    \hat{\mathbf{f}}^{\text{morph}}_i =
    \frac{\mathbf{f}^{\text{morph}}_i}{\lVert \mathbf{f}^{\text{morph}}_i \rVert_2},
    \qquad
    \hat{\mathbf{f}}^{\text{color}}_i =
    \frac{\mathbf{f}^{\text{color}}_i}{\lVert \mathbf{f}^{\text{color}}_i \rVert_2}.
\end{equation}
This normalization ensures that both descriptors lie on the unit hypersphere, so that their relative influence is governed by the fusion weights rather than by their original dimensionality.

An early fusion strategy was then adopted, in which the normalized descriptors were weighted and concatenated into a single feature vector prior to clustering:
\begin{equation}
    \mathbf{v}_i =
    \left[\, \alpha \, \hat{\mathbf{f}}^{\text{morph}}_i \,\|\,
    (1-\alpha) \, \hat{\mathbf{f}}^{\text{color}}_i \,\right]
    \in \mathbb{R}^{448},
\end{equation}
where $\alpha \in [0,1]$ controls the relative contribution of the morphological features and $1-\alpha$ that of the color features. Because both components are unit-normalized, the squared Euclidean distance between two fused vectors decomposes as
\begin{equation}
\begin{split}
    \lVert \mathbf{v}_i - \mathbf{v}_j \rVert_2^2 ={}&
    \alpha^2 \, \lVert \hat{\mathbf{f}}^{\text{morph}}_i - \hat{\mathbf{f}}^{\text{morph}}_j \rVert_2^2 \\
    &+ (1-\alpha)^2 \, \lVert \hat{\mathbf{f}}^{\text{color}}_i - \hat{\mathbf{f}}^{\text{color}}_j \rVert_2^2,
\end{split}
\end{equation}
so that $\alpha$ directly regulates the trade-off between structural and chromatic similarity. Three weighting schemes were evaluated, with $\alpha \in \{0.35, 0.50, 0.65\}$, in order to analyze the effect of this balance on the resulting cluster structure.

\subsubsection{Clustering and Evaluation}

Distance-based clustering algorithms such as K-Means tend to degrade in high-dimensional spaces, where pairwise distances become less discriminative. Therefore, the fused vectors were first projected onto a lower-dimensional manifold using Uniform Manifold Approximation and Projection (UMAP)~\cite{mcinnes2018umap}, a nonlinear dimensionality reduction technique that preserves the local neighborhood structure of the data while retaining much of its global organization. 

K-Means clustering~\cite{lloyd1982kmeans} was subsequently applied to the reduced representations for each number of clusters $K \in \{2, \ldots, 30\}$. The optimal value of $K$ was selected by jointly considering two internal validation criteria. The Elbow Method examines the within-cluster sum of squares as a function of $K$ and identifies the point beyond which additional clusters yield diminishing reductions. The Silhouette Score~\cite{rousseeuw1987silhouettes} measures, for each sample, how similar it is to its own cluster relative to the nearest neighboring cluster, ranging from $-1$ to $1$, where higher values indicate compact and well-separated clusters. This procedure was repeated independently for each value of $\alpha$.

\section{Results}
\label{sec:results}

The objective of the proposed method is to cluster the bus images and to determine whether the buses operated by RTP can be isolated into a coherent group without supervision. Accordingly, this section evaluates the proposed unsupervised early fusion approach against a baseline that clusters the images using morphological features alone. In addition, an ablation study was conducted to analyze the effect of the fusion weight $\alpha$ on clustering performance, considering three configurations: $\alpha = 0.65$, $\alpha = 0.50$, and $\alpha = 0.35$. The resulting clusters were assessed both quantitatively, by comparing them against the manually annotated target class, and qualitatively, through visual inspection of their contents.

\subsection{Implementation Details}

All components were implemented in Python and executed on a workstation equipped with an Intel Core i7 processor and an NVIDIA GTX 1080 GPU. To facilitate experimentation, feature extraction and clustering were implemented as independent stages. Morphological features were extracted with PyTorch~\cite{paszke2019pytorch} and Hugging Face Transformers~\cite{wolf2020transformers} using the pretrained \texttt{facebook/dinov2-small} checkpoint. Images were loaded with PIL, processed by the library's default image processor in batches of 64, and the resulting class tokens were stored as NumPy arrays. Color histograms were computed with OpenCV~\cite{bradski2000opencv} on the central crop of each image. Both descriptors were L2-normalized with scikit-learn~\cite{pedregosa2011scikit}, the fused matrix was embedded with the \texttt{umap-learn} library~\cite{mcinnes2018umap}, and K-Means clustering and Silhouette analysis were performed with scikit-learn.

\subsection{Evaluation Protocol}

For each configuration, Table~\ref{tab:results} reports the number of clusters $K$ selected by the Silhouette and Elbow criteria, together with the Shannon entropy of the distribution of RTP buses across clusters, defined as
\begin{equation}
    H = -\sum_{k=1}^{K} p_k \log_2 p_k,
\end{equation}
where $p_k$ denotes the fraction of all ground-truth RTP buses assigned to cluster $k$. The entropy equals zero when all RTP buses are grouped into a single cluster and increases as they become dispersed across multiple clusters. In addition, accuracy, precision, recall, and F1-score were computed by treating the cluster containing the largest number of RTP buses as the predicted RTP class and comparing its members against the manually annotated labels.

\begin{table}[tb]
\caption{Clustering results with respect to the target class (\textit{is RTP bus}) for different methods and values of $\alpha$. M.C.: morphological clustering (DINOv2 features only). E.F.: early fusion. Entropy (H) is reported in bits. Best values in bold.}
\label{tab:results}
\centering
\setlength{\tabcolsep}{3pt}
\begin{tabular}{lcccccc}
\hline
\textbf{Method} & $\boldsymbol{K}$ & \textbf{H} & \textbf{Acc.} & \textbf{Prec.} & \textbf{Rec.} & \textbf{F1} \\
\hline
M.C.                   & 11 & 1.175          & 0.9326          & 0.7703          & 0.6022          & 0.6759 \\
\hline
E.F. ($\alpha = 0.35$) & 3  & \textbf{0.012} & 0.6692          & 0.2606          & \textbf{0.9989} & 0.4133 \\
E.F. ($\alpha = 0.50$) & 20 & 1.680          & 0.9328          & 0.8988          & 0.4774          & 0.6236 \\
E.F. ($\alpha = 0.65$) & 14 & 1.059          & \textbf{0.9521} & \textbf{0.9507} & 0.6215          & \textbf{0.7516} \\
\hline
\end{tabular}
\end{table}

\subsection{Quantitative Analysis}

The early fusion configuration with $\alpha = 0.65$ achieved the best overall performance, with the highest accuracy (0.9521), precision (0.9507), and F1-score (0.7516) among all methods. Compared with the morphological baseline, incorporating color information increased precision by 18 percentage points (from 0.7703 to 0.9507) and the F1-score by 7.6 points (from 0.6759 to 0.7516), while also reducing the entropy from 1.175 to 1.059 bits. In contrast, recall improved only marginally (from 0.6022 to 0.6215). This indicates that the main contribution of the color descriptor is to increase cluster purity: the dominant RTP cluster obtained through fusion contains far fewer non-RTP vehicles, most likely buses from other routes that share a similar body structure but differ in livery, which morphological features alone cannot reliably distinguish.

Nevertheless, the recall of approximately 0.62 obtained by both the baseline and the best fusion configuration shows that a substantial fraction of RTP buses (about 38\%) is assigned to clusters other than the dominant one. This is consistent with the nonzero entropy values and suggests that the RTP class is not homogeneous in the feature space, possibly due to variations in viewpoint, illumination, partial occlusions, or differences among RTP vehicle models and paint schemes.

Performance did not vary monotonically with the color weight. With $\alpha = 0.50$, precision remained high (0.8988), but recall dropped to 0.4774 and the entropy reached its highest value (1.680 bits), indicating that the RTP buses were fragmented across a larger number of clusters ($K = 20$). As a result, the F1-score of this configuration (0.6236) fell below that of the morphological baseline. It should be noted that, because both descriptors are unit-normalized and the weights act quadratically on the squared Euclidean distance, $\alpha = 0.50$ assigns equal influence to both descriptors, whereas $\alpha = 0.35$ gives color roughly 3.4 times the influence of morphology.

The configuration with $\alpha = 0.35$ illustrates the failure mode of an excessive color weight. Although it yielded the lowest entropy (0.012 bits) and a near-perfect recall (0.9989), these values do not reflect a successful separation. With only three clusters, virtually all RTP buses fall into a single cluster, but this cluster also contains a large number of non-RTP vehicles, as reflected by a precision of only 0.2606 and the lowest accuracy and F1-score of all configurations (0.6692 and 0.4133, respectively). In other words, the low entropy is a consequence of the collapse of the clustering space rather than of an accurate grouping of the target class. This case highlights that entropy and recall must be interpreted jointly with precision, since each of them can be trivially optimized by a degenerate solution.

Accuracy should also be interpreted with caution. Because RTP buses represent a minority of the dataset, a trivial classifier that labels every image as non-RTP would already achieve a high accuracy. The F1-score and precision are therefore more informative indicators of the quality of the RTP cluster.

Overall, these results show that, with an appropriate fusion weight, the proposed methodology groups the majority of the target class into a single, highly pure cluster. This supports our hypothesis and suggests that the approach can facilitate the detection of public transport vehicles without the need for large annotated datasets.

\subsection{Qualitative Analysis}

\begin{figure}[tb]
\centering
    \includegraphics[width=0.23\columnwidth,height=0.18\columnwidth,keepaspectratio=false]{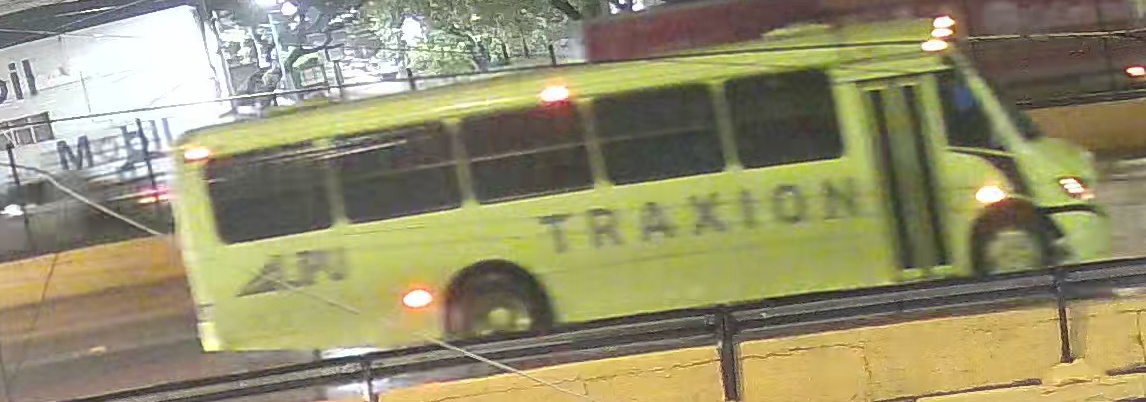}\hfill
    \includegraphics[width=0.23\columnwidth,height=0.18\columnwidth,keepaspectratio=false]{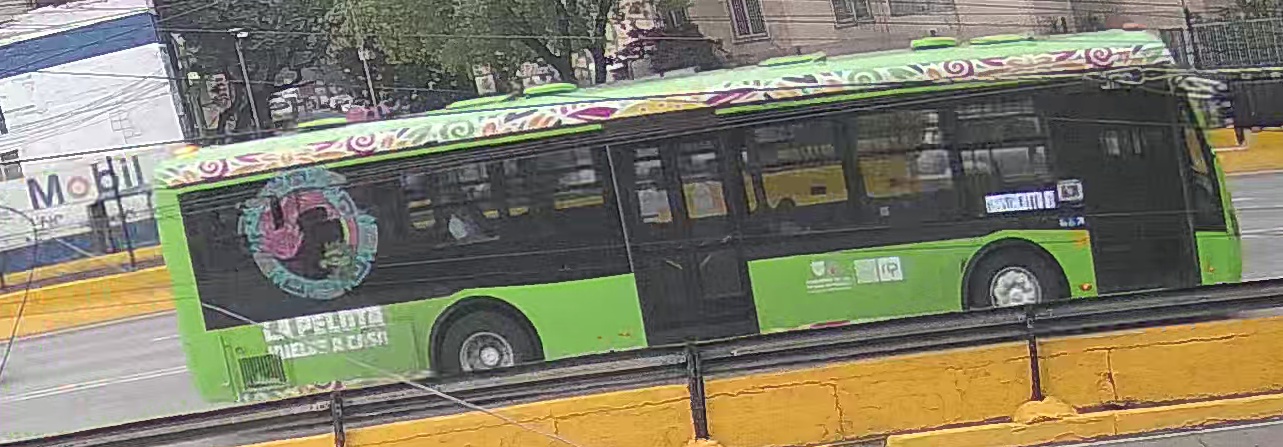}\hfill
    \includegraphics[width=0.23\columnwidth,height=0.18\columnwidth,keepaspectratio=false]{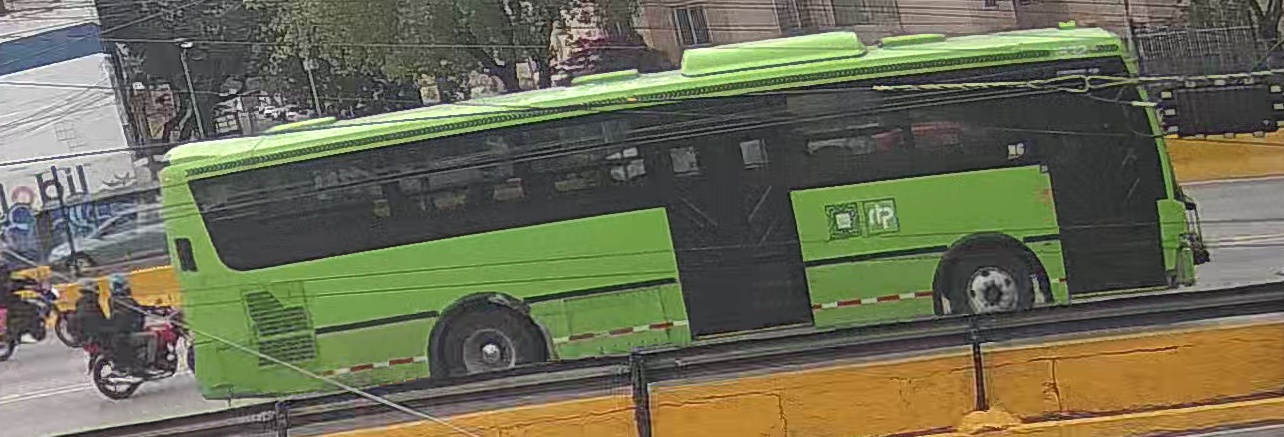}\hfill
    \includegraphics[width=0.23\columnwidth,height=0.18\columnwidth,keepaspectratio=false]{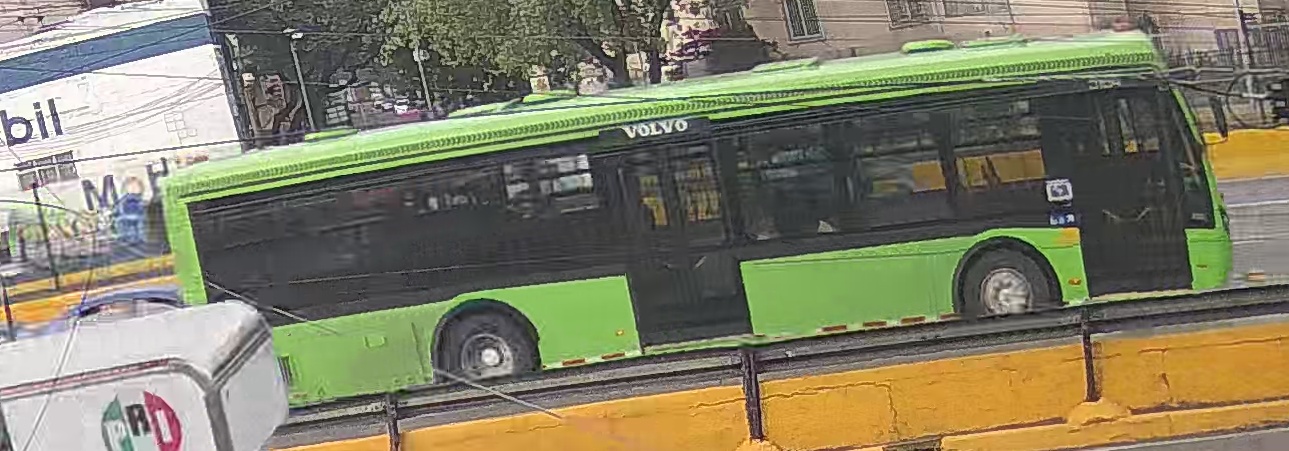}\\[4pt]
    \includegraphics[width=0.23\columnwidth,height=0.18\columnwidth,keepaspectratio=false]{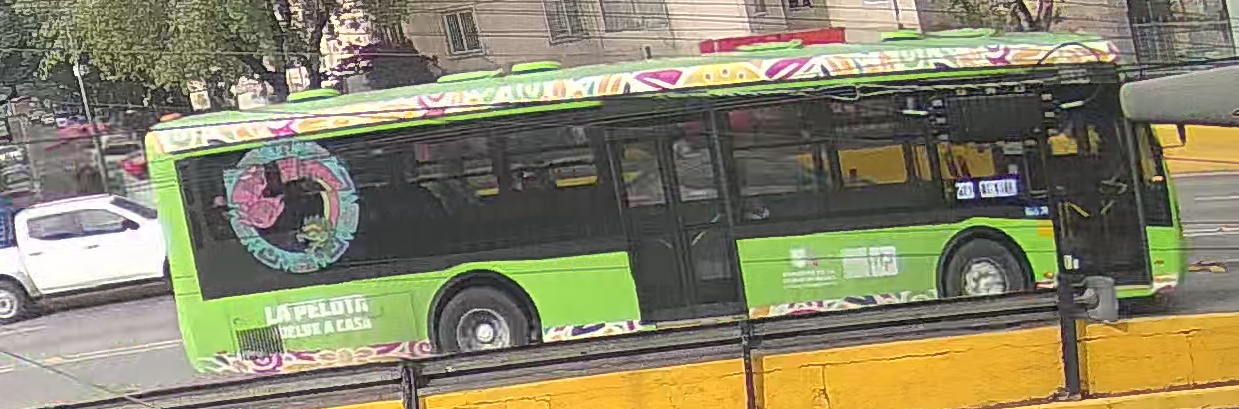}\hfill
    \includegraphics[width=0.23\columnwidth,height=0.18\columnwidth,keepaspectratio=false]{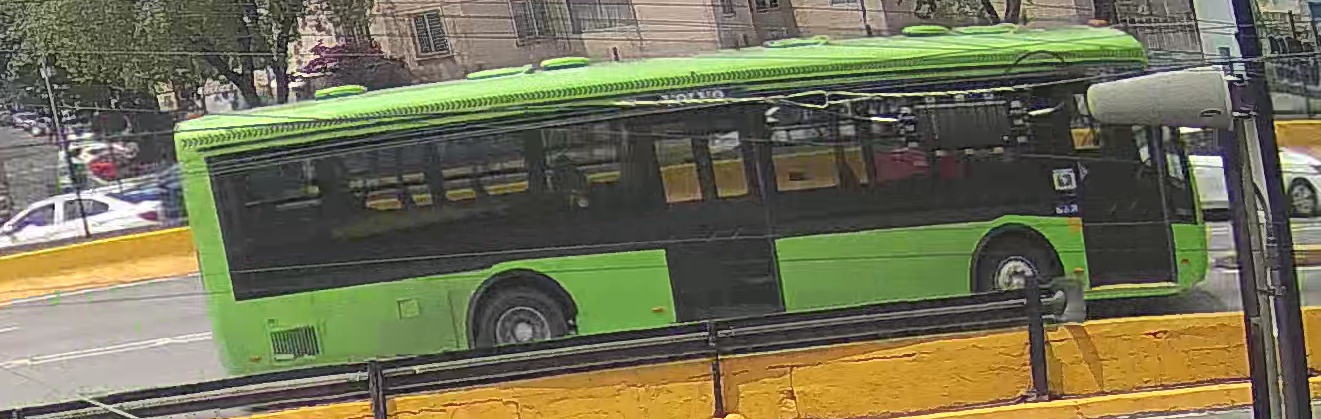}\hfill
    \includegraphics[width=0.23\columnwidth,height=0.18\columnwidth,keepaspectratio=false]{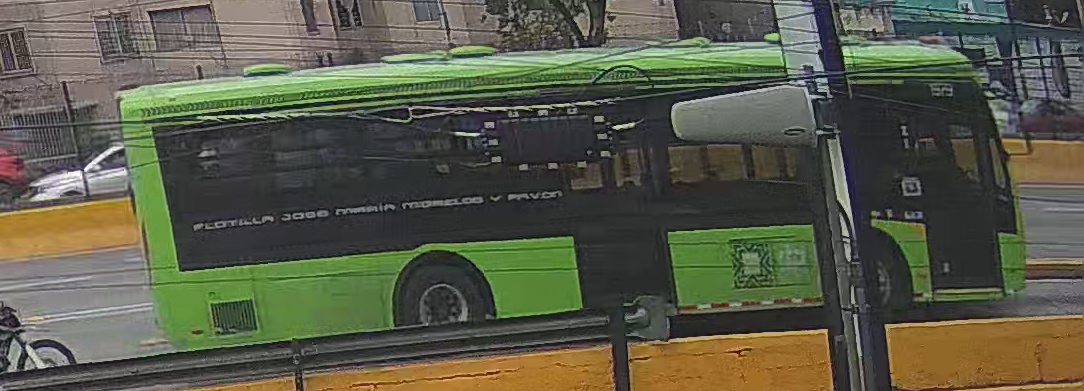}\hfill
    \includegraphics[width=0.23\columnwidth,height=0.18\columnwidth,keepaspectratio=false]{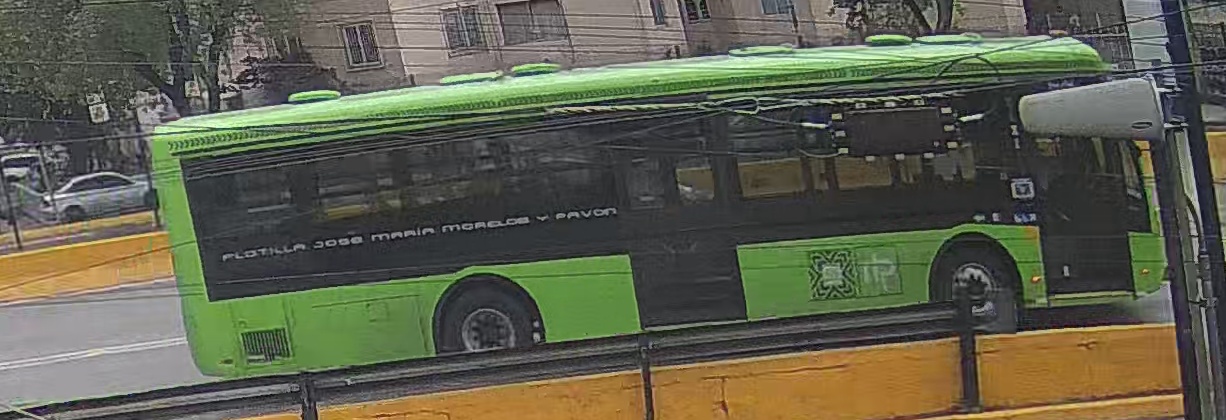}\\[4pt]
    \includegraphics[width=0.23\columnwidth,height=0.18\columnwidth,keepaspectratio=false]{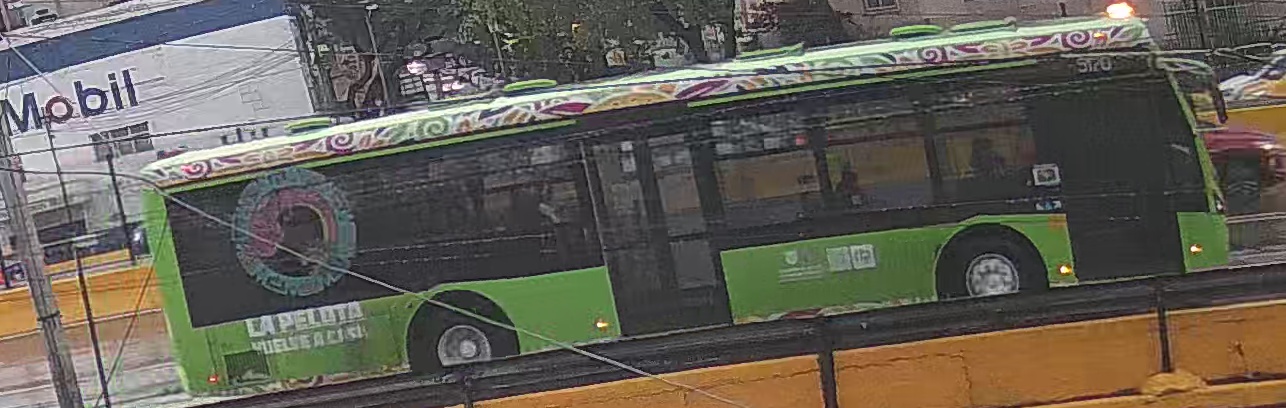}\hfill
    \includegraphics[width=0.23\columnwidth,height=0.18\columnwidth,keepaspectratio=false]{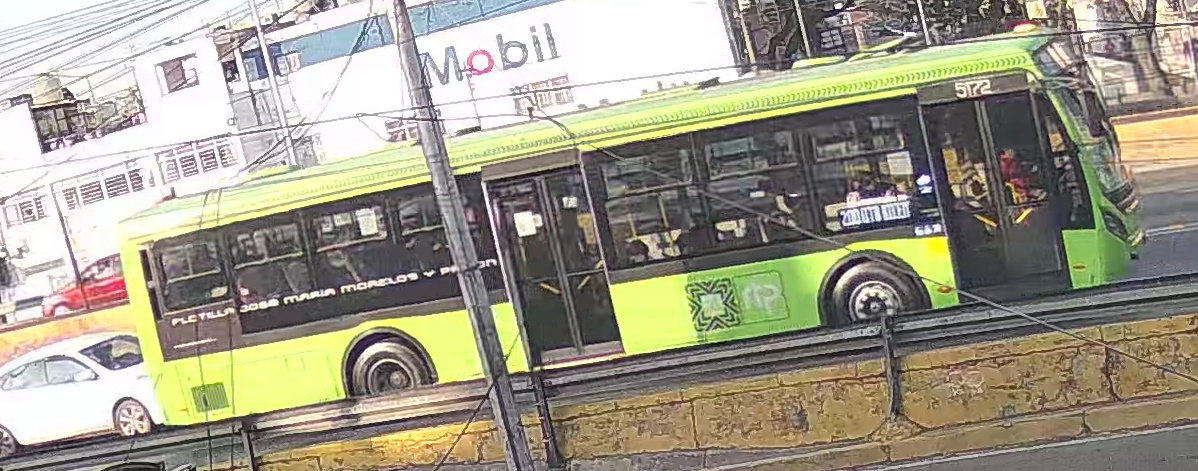}\hfill
    \includegraphics[width=0.23\columnwidth,height=0.18\columnwidth,keepaspectratio=false]{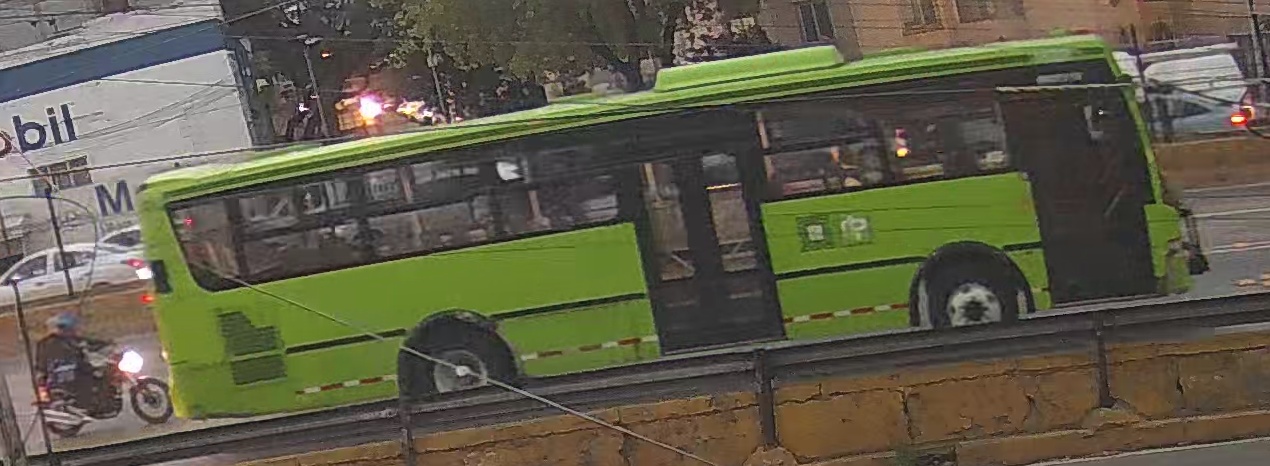}\hfill
    \includegraphics[width=0.23\columnwidth,height=0.18\columnwidth,keepaspectratio=false]{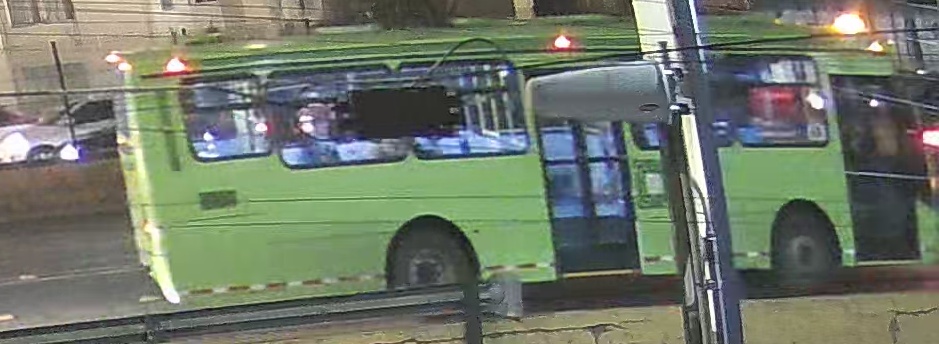}
\caption{Randomly selected images from the dominant RTP cluster obtained with early fusion ($\alpha = 0.65$).}
\label{fig:rtp_cluster}
\end{figure}

Visual inspection of the resulting clusters supports the quantitative findings. With $\alpha = 0.65$, the Silhouette analysis indicated an optimal value of $K = 14$, and the clusters were highly homogeneous: vehicles with distinct shapes were correctly separated, while buses belonging to the same transit route remained chromatically consistent within their clusters. Fig.~\ref{fig:rtp_cluster} shows twelve randomly selected members of the dominant RTP cluster.

Increasing the color weight to $\alpha = 0.50$ shifted the optimal number of clusters to $K = 20$. In this configuration, cohesive groups were split into fragmented sub-clusters that appeared to be driven primarily by minor variations in lighting or saturation rather than by differences in vehicle structure.

When color dominated the fusion ($\alpha = 0.35$), the clustering collapsed to $K = 3$. The model became largely insensitive to structural differences, grouping morphologically distinct vehicles, such as vans and full-size buses, into the same cluster simply because they shared dominant colors.

Overall, these results confirm that color is a highly discriminative cue for identifying public transport routes, but that it must complement, rather than dominate, the morphological representation. A moderate color weight improves the purity of the target cluster, whereas over-prioritizing color degrades the structural understanding of the vehicles and leads either to over-segmentation or to the collapse of the clustering space.

\section{Conclusions}
\label{sec:conclusions}

This study presented an unsupervised pipeline for clustering public transport vehicles captured by a surveillance camera, in which frozen DINOv2 morphological features and HSV color histograms are combined through weighted early fusion and grouped with UMAP and K-Means. Without using any labels during feature extraction or clustering, the best configuration isolated RTP buses in a cluster with higher precision and F1-score than morphological features alone. These results confirm that color provides information that self-supervised features alone do not capture well, since it separates buses that share a body structure but belong to different routes.

The ablation study also showed that the fusion weight is critical. A balanced weight fragmented the target class into many small clusters, whereas a color-dominant weight collapsed the space into three clusters that mixed vans, microbuses, and full-size buses. Color should therefore complement, not replace, the morphological representation. 

The approach has limitations. Some RTP buses were still assigned to secondary clusters and the number of clusters was selected with internal criteria that do not guarantee semantic correspondence. Future work will explore late fusion based on descriptor-specific distances combined with density-based algorithms such as HDBSCAN, which do not require the number of clusters to be fixed in advance. Evaluation on multiple cameras and routes will also be necessary to assess the generality of the method.

\bibliographystyle{IEEEtran}
\bibliography{referencias}

\end{document}